\pdfoutput=1

\documentclass[11pt]{article}

\usepackage[dvipsnames,table,x11names]{xcolor}

\usepackage[final]{acl}

\usepackage{times}
\usepackage{latexsym}

\usepackage[T1]{fontenc}

\usepackage[utf8]{inputenc}

\usepackage{microtype}

\usepackage{inconsolata}

\usepackage{graphicx}

\usepackage{microtype}      
\usepackage{amsmath}
\usepackage{graphicx}
\usepackage{algorithm}
\usepackage{algorithmic}
\usepackage{array} 
\usepackage{soul}
\usepackage{tcolorbox}
\usepackage{caption}
\usepackage{newfloat}
\usepackage{changepage} 
\usepackage{amssymb}

\usepackage{colortbl}
\DeclareFloatingEnvironment[fileext=lop]{Prompt}

\title{Fair Summarization: Bridging Quality and Diversity in Extractive Summaries}

\author{Sina Bagheri Nezhad, Sayan Bandyapadhyay, Ameeta Agrawal \\
  Department of Computer Science \\
  Portland State University, USA \\
  \texttt{\{sina5,sayanb,ameeta}@pdx.edu\}}

\begin{document}
\maketitle
\begin{abstract}
Fairness in multi-document summarization of user-generated content remains a critical challenge in natural language processing (NLP). Existing summarization methods often fail to ensure equitable representation across different social groups, leading to biased outputs. In this paper, we introduce two novel methods for fair extractive summarization: \texttt{FairExtract}, a clustering-based approach, and \texttt{FairGPT}, which leverages GPT-3.5-turbo with fairness constraints. We evaluate these methods using \texttt{Divsumm} summarization dataset of White-aligned, Hispanic, and African-American dialect tweets and compare them against relevant baselines. The results obtained using  a comprehensive set of summarization quality metrics such as SUPERT, BLANC, SummaQA, BARTScore, and UniEval, as well as a fairness metric \(F\),  demonstrate that \texttt{FairExtract} and \texttt{FairGPT} achieve superior fairness while maintaining competitive summarization quality. Additionally, we introduce composite metrics (e.g., SUPERT+\(F\), BLANC+\(F\)) that integrate quality and fairness into a single evaluation framework, offering a more nuanced understanding of the trade-offs between these objectives. Our code is available online.\footnote{\url{https://github.com/PortNLP/FairEXTSummarizer}} 
\end{abstract}

\section{Introduction}

\begin{table*}[!t]
\centering
\definecolor{group1}{RGB}{255, 200, 200} 
\definecolor{group2}{RGB}{200, 200, 255} 
\small
\begin{tabular}{|p{7.5cm}|p{7.5cm}|} 
\hline
\textbf{ChatGPT-EXT \citep{zhang-etal-2023-extractive-summarization}} & \textbf{FairGPT (Ours)} \\ \hline
\sethlcolor{group1}
\hl{If you see on the news something about the Chicago Kitchen Clown Bandits then it will be referring me my friend Eten and I.} 
\sethlcolor{group1}
\hl{Turns out not all White Castles are the same. Why do you push me away Chicago?!} 
\sethlcolor{group2}
\hl{I mean I'm from Chicago. I'll cheer for the Bears, but I'm a bigger 49ers fan.} 
\sethlcolor{group2}
\hl{Is this new wave of Chicago Rap gonna be like the Hyphy movement?} 
\sethlcolor{group1}
\hl{Don't talk shot about Chicago, or those big shoulders will plow right into your little Boston ass.} 
\sethlcolor{group1}
\hl{Nothing makes me happier than seeing the Bulls win \#ChicagoBasketball \#Bullieve.} 
& 
\sethlcolor{group1}
\hl{Don't talk shot about Chicago, or those big shoulders will plow right into your little Boston ass.} 
\sethlcolor{group1}
\hl{Nothing makes me happier than seeing the Bulls win \#ChicagoBasketball \#Bullieve.} 
\sethlcolor{group2}
\hl{Truuu we tryna find sum to do too.. I dnt wanna b n Chicago if ain't nobody here.} 
\sethlcolor{group1}
\hl{Turns out not all White Castles are the same. Why do you push me away Chicago?!} 
\sethlcolor{group2}
\hl{I mean I'm from Chicago. I'll cheer for the Bears, but I'm a bigger 49ers fan.} 
\sethlcolor{group2}
\hl{Is this new wave of Chicago Rap gonna be like the Hyphy movement?} \\ \hline
\end{tabular}
\caption{Comparison of summaries generated by ChatGPT-EXT and FairGPT. Tweets from different groups are highlighted: \colorbox{group1}{Group 1 (e.g., White-aligned)} and \colorbox{group2}{Group 2 (e.g., African-American)}.}
\label{tab:summary_comparison}
\end{table*}

Multi-document summarization, which condenses multiple documents into a concise summary, is a fundamental task in natural language processing (NLP). Summarization methods are typically either \textit{extractive}, selecting the most important sentences, or \textit{abstractive}, where the content is rephrased.

Early research focused on summarizing formal text sources such as news articles. However, with the rise of social media, attention has shifted to summarizing user-generated content, which is diverse in style and language \citep{dash2019summarizing,jung-etal-2019-earlier,10.1145/3442381.3450108,olabisi-etal-2022-analyzing}. Social media platforms bring together users from varied backgrounds, introducing linguistic diversity through informal language, slang, and emojis. This diversity raises the challenge of ensuring fairness in summarization for a balanced representation of various social groups. In social media, where public opinion is shaped, fair summaries are essential to include different perspectives and avoid underrepresentation of one or more social groups as without proper representation, certain voices might be excluded or misrepresented. 
Therefore, ensuring that all groups—across race, gender, and linguistic diversity—are fairly represented is critical for generating balanced summaries that reflect the diversity of public opinion \citep{Dash2018SummarizingUT}. 
In particular, the dialectal variations among White-aligned, Hispanic, and African-American groups not only reflect different linguistic styles but also embody distinct cultural expressions that influence how users communicate.

Despite advancements, bias remains a concern in automated summarization \citep{dash2019summarizing,jung-etal-2019-earlier,10.1145/3442381.3450108,olabisi-etal-2022-analyzing} as most existing summarization methods focus on quality but fall short in optimizing fairness. Improving fairness can sometimes lower quality \citep{jung-etal-2019-earlier}. This gap leads to the key question: if a summarization method is optimized for fairness, how does it affect the overall summary quality?


In this paper, we address two research questions:
\begin{enumerate}
    \item How does achieving perfectly fair summaries affect overall quality?
    \item How well do current methods perform when considering both fairness and quality?
\end{enumerate}

To illustrate the performance of fairness-aware summarization models, we compare summaries generated by ChatGPT-EXT \citep{zhang-etal-2023-extractive-summarization} and our proposed FairGPT model on a sample instance from \texttt{Divsumm} dataset \citep{olabisi-etal-2022-analyzing}. As shown in Table \ref{tab:summary_comparison}, FairGPT ensures equal representation of tweets from different groups, while ChatGPT-EXT shows a slight imbalance. 

We make the following contributions:
\begin{itemize}
    \item We propose \texttt{FairExtract}, a fair clustering-based extractive summarization method that achieves perfect fairness while preserving competitive summarization quality, as demonstrated through evaluations against baseline models using standard and composite quality-fairness metrics.
    \item We develop \texttt{FairGPT}, a large language model-based extractive summarization method that enforces fairness through equal representation and accurate content extraction using the longest common subsequence, producing fair summaries without sacrificing competitive summarization quality.
    \item We introduce composite metrics combining normalized quality scores with fairness, providing a comprehensive analysis of the quality-fairness trade-off in summarization models.
\end{itemize}

\section{Related Work}

The field of NLP has increasingly focused on addressing bias and fairness, with research focused along two key dimensions: intrinsic bias, stemming from text representations, and extrinsic bias, reflecting performance disparities across demographic groups \citep{han-etal-2023-fair}.

Early work on fairness in summarization \citep{10.1145/3184558.3186947, dash2019summarizing} revealed that summaries often fail to represent source data fairly, even when source texts from different groups have similar quality. This led to the development of fairness-aware algorithms across various stages of summarization, including pre-processing, in-processing, and post-processing techniques. For example, \citet{10.1145/3442381.3450108} proposed a post-processing method to mitigate dialect-based biases. \citet{olabisi-etal-2022-analyzing} introduced the DivSumm dataset, focusing on dialect diversity in summarization and evaluating algorithms on fairness.

Recent work has explored bias related to the position of input data. \citet{olabisi-agrawal-2024-understanding} studied position bias in multi-document summarization, showing that the order of input texts affects fairness. Similarly, \citet{huang-etal-2023-examining} analyzed clustering-based summarization models, which may introduce political or opinion bias, emphasizing the need for fair representation.

Recent work highlights that large language models often reflect dominant Western cultural norms, resulting in cultural bias \cite{10.1093/pnasnexus/pgae346}. \citet{liu2024culturallyawareadaptednlp} provided a taxonomy for culturally aware NLP that emphasizes the role of values, norms, and linguistic diversity. Moreover, \citet{hershcovich-etal-2022-challenges} discussed cross-cultural challenges in NLP and advocate for strategies that integrate cultural insights into model development. 

Fair clustering, another key technique, has also seen significant research. \citet{NIPS2017_978fce5b} introduced the concept of fairlets—small, balanced clusters that ensure fair representation across protected groups. Building on this, \citet{pmlr-v97-chen19d} proposed proportional centroid clustering to eliminate biases in cluster-based models.

Further advancements include scalable techniques for fair clustering, such as the fair $k$-median clustering method \citep{backurs2019scalable}, and approaches that generalize fairness constraints across multiple protected groups \citep{NEURIPS2019_fc192b0c}. \citet{NEURIPS2020_95f2b84d} extended this work to probabilistic fair clustering, offering solutions for uncertain group memberships.

In the domain of clustering methodologies, \citet{micha_et_al:LIPIcs:2020:12492} explored fairness in centroid clustering, while \citet{Li_2020_CVPR} proposed Deep Fair Clustering (DFC), which leverages deep learning to filter sensitive attributes, improving both fairness and performance. This underscores the growing importance of combining fairness with robust clustering methods in NLP tasks.

\section{Task Formulation}
In this work, we address the challenge of diversity-preserving multi-document extractive summarization. Given a collection of documents $\mathcal{D} = \{d_1, d_2, \ldots, d_n\}$ from two diverse social groups, $G_1$ and $G_2$, the goal is to produce an extractive summary $\mathcal{S} = \{s_1, s_2, \ldots, s_k\} \subset \mathcal{D}$ of length $k << n$, ensuring balanced representation from both groups.

In this context, each document is a tweet from a specific dialect group, which serves as an indicator of its social group. Traditionally, various metrics like ROUGE \citep{lin-2004-rouge} and BERTScore \citep{zhang2019bertscore} have been used to evaluate summary quality. However, our primary focus is on balancing both quality and fairness, particularly in terms of representing different social groups equitably. To measure fairness, we use the \textit{Representation Gap (RG)} metric, as proposed by \citet{olabisi-etal-2022-analyzing}. This metric captures how well the summary reflects the proportions of the original groups. A lower RG score indicates better balance and thus a fairer summary.

For a summary $\mathcal{S}$ of length $k$, let $N_1(\mathcal{S})$ and $N_2(\mathcal{S})$ represent the number of documents from groups $G_1$ and $G_2$, respectively. The Representation Gap is defined as:
\begin{equation}
\text{RG}(\mathcal{S}) = 
\textstyle\frac{\max\{N_1(\mathcal{S}), N_2(\mathcal{S})\} - 
\min\{N_1(\mathcal{S}), N_2(\mathcal{S})\}}{k}.
\end{equation}

For example, if $k = 6$, with 4 documents from $G_1$ and 2 from $G_2$, the RG is 0.333. When both groups are equally represented, the RG is 0, indicating a \emph{perfectly fair} summary.

At this point, we recognize two key challenges: 
(1) While quality metrics improve with larger values, fairness improves with smaller Representation Gap (RG) values.
(2) Quality and fairness metrics differ greatly in scale, making direct comparison difficult.

To address these issues, we introduce a new fairness metric, \( F \), defined as:

\begin{equation}  
F(\mathcal{S}) = 1 - \text{RG}(\mathcal{S})
\end{equation}

This transformation ensures that larger \( F \) values indicate better fairness, aligning it with the behavior of quality metrics. Furthermore, we apply min-max normalization to rescale all metrics to the range \([0, 1]\), ensuring comparability across different scales. The normalization formula is given by:

\begin{equation}
\frac{\text{value} - \min}{\max - \min}
\end{equation}

where $\min$ and $\max$ are the minimum and maximum observed values for the respective metric.

Finally, we introduce composite metrics, such as \textbf{SUPERT+F}, \textbf{BLANC+F}, \textbf{SummaQA+F}, \textbf{BARTScore+F}, and \textbf{UniEval+F}, which are the averages of the normalized quality metrics (e.g., SUPERT \citep{gao-etal-2020-supert}, BLANC \citep{vasilyev-etal-2020-fill}, SummaQA \citep{scialom-etal-2019-answers}, BARTScore \citep{yuan2021bartscore}, and UniEval \citep{zhong-etal-2022-towards}) and the fairness score \(F\), providing a balanced assessment of both quality and fairness. 

\section{Fair Extractive Summarizers}
In this work, we introduce two novel methods for fair extractive summarization: FairExtract and FairGPT. FairExtract utilizes clustering techniques with fairlet decomposition to ensure diversity in summaries while maintaining high-quality representation across different groups. FairGPT, on the other hand, leverages large language models (LLMs) such as GPT-3.5, incorporating fairness constraints and the longest common subsequence (LCS) method to match and fairly select content from different groups. Both methods prioritize fairness and ensure equitable representation in the generated summaries.

\subsection{FairExtract: A Clustering-based Fair Extractive Summarization Method}

The task of clustering is central to the FairExtract process, which aims to generate diversity-preserving summaries. The method combines document embeddings, fairlet decomposition, and clustering techniques to ensure both fairness and quality. Below, we describe the steps involved in detail:

\begin{enumerate}
    \item \textbf{Embedding Documents:}
    We begin by embedding each document (tweet) into a high-dimensional space (e.g., using a pretrained model such as BERT \citep{devlin-etal-2019-bert}), capturing its semantic content in Euclidean space. This embedding enables us to compute meaningful distances between documents, which is crucial for clustering.
    \item \textbf{Fairlet Decomposition:}
    To ensure fairness in the summarization process, we decompose the dataset into fairlets. A fairlet is the smallest set of documents that maintains proportional balance between two groups, $G_1$ and $G_2$ \citep{backurs2019scalable}. Assume the desired ratio of documents from $G_1$ to $G_2$ is $g_1:g_2$, where $g_1$ and $g_2$ are coprime (i.e., $\gcd(g_1, g_2) = 1$). Then, a fairlet is defined as the smallest group of documents that exactly preserves this ratio, containing exactly $g_1$ documents from $G_1$ and $g_2$ documents from $G_2$. This ensures that the composition of the fairlet reflects the required ratio between the two groups, maintaining fairness at the smallest possible scale. The decomposition aims to minimize the sum of Euclidean distances between documents within the same fairlet.
    \item \textbf{Finding the Fairlet Center:}
    Once the dataset is divided into fairlets, we compute the center of each fairlet. The center is the document within the fairlet that minimizes the sum of distances to all other documents in the same fairlet. This document acts as the representative of the fairlet, summarizing the content while maintaining group balance.
    \item \textbf{$k$-Median Clustering on Fairlet Centers:}
    After identifying the centers of all fairlets, we apply the $k$-median clustering algorithm to these centers. In the $k$-median problem, we are given a set of points $P$ in a $d$-dimensional space, and we aim to partition them into $k$ clusters $\Pi = \{P_1, \ldots, P_k\}$ that minimize the following cost:
    \begin{equation}
        \min_{C\subset P: |C|=k} \sum_{c_i\in C\mid 1\le i\le k} \sum_{p\in P_i} ||p-c_i||.
    \end{equation}
    The number of clusters $k$ is selected such that $k \times (g_1 + g_2)$ equals the desired number of documents in the summary. This step ensures that the clusters formed are representative of both social groups.
    \item \textbf{Summary Construction:}
     From each $k$-median cluster, we select the center fairlet and include all documents within that fairlet in the final summary. By selecting one fairlet from each cluster, we maintain both quality and fairness, as the summary reflects the balanced representation of both groups. The resulting extractive summary ensures that the most salient information is captured while maintaining equitable representation of the social groups.
\end{enumerate}

For a formal representation of the process, see Appendix \ref{sec:appendix-algo}. 

\subsection{FairGPT: An LLM-based Fair Extractive Summarization Method}

\begin{figure*}[t] 
\small
\begin{adjustwidth}{0in}{-1in} 
\begin{tcolorbox}[colback=blue!5!white, colframe=gray!70!black, title=FairGPT Prompt, width=\textwidth]
\begin{verbatim}
system:  "You are an extractive fair summarizer that follows the output pattern. A fair summarizer
          should select the same number of sentences from each group of people." 
             
user:    "Please extract sentences as the summary. The summary should contain {L} sentences
          which means select {L/2} number of sentences from each group of people to represent
          the idea of all groups in a fair manner.  
          Document:{document}"
\end{verbatim}
\end{tcolorbox}
\end{adjustwidth}
\vspace{-10pt}
\captionof{Prompt}{Prompt used in FairGPT. The variable \texttt{L} refers to the total number of sentences to be extracted.}
\label{promptbox}
\end{figure*}

\begin{algorithm*}[!ht]
\caption{FairGPT Algorithm}
\small
\label{alg:fairgpt}
\begin{algorithmic}
    \STATE \textbf{Input:}
    \begin{itemize}
        \item Document set $\mathcal{D}$ divided into groups $G_1$ and $G_2$
        \item Desired summary length \(L\) with \(L/2\) sentences from each group
    \end{itemize}

    \STATE \textbf{Output:} Fair extractive summary $\mathcal{S}$
    \vspace{0.2cm}
    \STATE \textbf{Step 1: Input Preparation} \\
    Create documents for $G_1$ and $G_2$, clearly labeling each sentence based on its group.
    \vspace{0.1cm}
    \STATE \textbf{Step 2: Summarization using LLM} \\
    Instruct LLM (GPT-3.5-turbo) using Prompt \ref{promptbox} to select \(L/2\) sentences from each group, ensuring fair representation.
    \vspace{0.1cm}
    \STATE \textbf{Step 3: Matching using Longest Common Subsequence (LCS)} \\
    Use LCS to match the GPT-generated sentences with the original dataset to identify the closest matching tweets and include the full sentences in the summary.
    \vspace{0.1cm}
    \STATE \textbf{Step 4: Ensuring 50\% Similarity} \\
    Ensure that at least 50\% of the content in each generated sentence matches the corresponding original tweet using LCS.
    \vspace{0.1cm}
    \STATE \textbf{Step 5: Fairness Check} \\
    Verify that the summary contains an equal number of sentences from $G_1$ and $G_2$. If fairness or similarity conditions are not met, go to Step 2.
    \vspace{0.1cm}
    \STATE \textbf{Step 6: Final Output} \\
    Save the final summary $\mathcal{S}$ once both fairness and quality thresholds are satisfied.
    \vspace{0.1cm}
    \STATE \textbf{Return:} The final summary $\mathcal{S}$.
\end{algorithmic}
\end{algorithm*}

FairGPT leverages an LLM generate fair extractive summaries by selecting an equal number of sentences from different social groups. It applies fairness checks and uses the longest common subsequence (LCS) to match generated summaries with the original tweets. Below are the detailed steps:

\begin{enumerate}
    \item \textbf{Input Preparation:}  
    The dataset is split into two groups (e.g., White-aligned and Hispanic dialects), and a document with sentences for each group is created as input for the summarization process.

    \item \textbf{Summarization using an LLM:}  
    We use an LLM (GPT-3.5-turbo) to generate a summary of length \(L\), selecting \(L/2\) sentences from each group to ensure balanced representation. The specific prompt used for this task is available in the Prompt \ref{promptbox}.

    \item \textbf{Matching using Longest Common Subsequence (LCS):}  
    As GPT sometimes extracts partial sentences, we apply LCS to match the generated summary with the closest original tweets. The full tweets corresponding to the longest common subsequences are added to the final summary.

    \item \textbf{Output Check:}  
    After generating the summary, we verify two key aspects. First, at least 50\% of the content in each GPT-generated sentence must match the corresponding original tweet using the LCS. Second, we ensure that the summary is perfectly fair, with equal representation from each group.

    This output check is crucial because large language models, such as GPT-3.5-turbo, sometimes generate unexpected outputs that do not align with the input instructions. To ensure the generated summaries meet both fairness and content similarity criteria, we repeat the process if either condition is not satisfied. In our tests of generating 75 summaries, the repetition process never exceeded 10 iterations, and the average number of repetitions across all tests was 1.6, indicating the efficiency and reliability of the output check mechanism.

    \item \textbf{Final Output:}  
    Once the summary satisfies both fairness and similarity requirements, it is saved as the final output.
\end{enumerate}

For a formal representation of the process, see Algorithm \ref{alg:fairgpt}.

\section{Experimental Setup}
Next, we describe the dataset, baseline methods, and evaluation metrics that are used to comprehensively assess the quality and fairness of the generated summaries.

\begin{table*}[!t]
\centering
\small
\renewcommand{\arraystretch}{1.4} 
\setlength{\tabcolsep}{10pt} 
\rowcolors{1}{gray!15}{white} 
\begin{tabular}{|>{\centering\arraybackslash}m{2.3cm}|>{\raggedright\arraybackslash}m{12cm}|}
\hline
\rowcolor{gray!30} \textbf{Group} & \textbf{Tweet} \\ \hline
White-aligned & Turns out not all White Castles are the same. Why do you push me away Chicago?! \\ \hline
African American & "I mean I'm from Chicago. I'll cheer for the Bears, but I'm a bigger 49ers fan." \\ \hline
White-aligned & Nothing makes me happier than seeing the Bulls win \_\_\_\_\_ \#ChicagoBasketball \#Bullieve \\ \hline
White-aligned & If you see on the news something about the Chicago Kitchen Clown Bandits, then it will be referring to me, my friend Eten, and I. \\ \hline
African American & Truuu we tryna find sum to do too.. I dnt wanna b n Chicago if ain't nobody here. \\ \hline
White-aligned & Oh yeah.. I'm good. Hangin' up here in Chicago today. :) \\ \hline
Hispanic & You girls have a safe flight.! See you in Chicago (: \\ \hline
\rowcolor{gray!15}
\textbf{...} & \textbf{... (Dataset continues with more examples)} \\ \hline
\end{tabular}
\caption{Sample of tweets from different social groups in the dataset. The full dataset contains many more examples.}
\label{tab:dataset-sample}
\end{table*}

\subsection{Dataset}
The dataset used in this study is \textit{DivSumm} \citep{olabisi-etal-2022-analyzing}, consisting of tweets from three dialect groups—White-aligned, Hispanic, and African-American—across 25 topics, with 30 tweets per group per topic, totaling 2,250 tweets. 

Our model works with two groups at a time, so we explore three pairings: White-Hispanic, Hispanic-African American, and White-African American. Each pairing maintains proportional representation from both groups to ensure an equitable balance in the summarization process. Table \ref{tab:dataset-sample} presents a sample of the dataset used in this study, containing tweets from different social groups about Chicago. 

For our experiments, we formed 60 tweets per group pair (30 from each group) and generated a 6-tweet summary per pair, covering all 25 topics. This yielded 75 distinct summaries per model, allowing us to evaluate both fairness and quality comprehensively. 

\subsection{Baseline Methods}
 
Here, we provide a detailed description of the baseline methods used in our comparative analysis:

\noindent \textbf{Naive:} In the Naive baseline method, $L$  tweets are randomly chosen from the input without any specific criteria. This approach represents a straightforward, non-strategic selection process and serves as a basic reference point for evaluating other methods.

\noindent \textbf{NaiveFair:} The NaiveFair baseline method involves randomly selecting $L/2$ tweets from each social group. This method ensures equal representation from each group, providing a basic notion of fairness without any sophisticated processing.

For the Naive and NaiveFair methods, which involve randomness in selecting summaries, we conducted the experiment five times for each summary, resulting in 375 different summaries for each of these methods.

\noindent \textbf{TextRank:} TextRank is an unsupervised graph-based ranking method used for extractive summarization \citep{mihalcea-tarau-2004-textrank}. This standard \texttt{vanilla} baseline approach uses a single aggregated set of randomized documents from all groups as input for summarization, without any pre-processing.

\noindent \textbf{BERT-Ext:} BERT-Ext is an extractive summarization model that utilizes pre-trained embeddings from BERT and k-means clustering to select sentences closest to the centroid as summaries \citep{miller2019leveraging}. Similar to the TextRank baseline, we implemented BERT-Ext \texttt{vanilla} method.

\noindent \textbf{Cluster-Heuristic (Cluster-H)}: This method first partitions the input documents into group-based subsets before generating separate group summaries of length $ $. These group-level summaries are shuffled, combined and then used to generate a final, unified summary \citep{dash2019summarizing,olabisi-etal-2022-analyzing}. As summarization models, we use TextRank and BERT-Ext. 

\noindent \textbf{Cluster-Automatic (Cluster-A)}: In this attribute-agnostic approach, documents are clustered automatically into $m$ subsets, and corresponding summaries of length $ $ are generated. The summaries are concatenated and used to generate a final summary \citep{olabisi-etal-2022-analyzing}. As summarization models, we experiment with TextRank and BERT-Ext. 

\noindent \textbf{ChatGPT-EXT}: This approach uses GPT-3.5 for extractive summarization by employing in-context learning and chain-of-thought reasoning to identify key sentences. It focuses on extracting salient content from documents to generate coherent summaries while maintaining the structure of the original text \citep{zhang-etal-2023-extractive-summarization}.

\subsection{Evaluation Metrics}

Below, we list the several reference-free metrics which do not rely on human-written reference text used for evaluation in this study.
\begin{itemize}
    \item \textbf{SUPERT:} SUPERT \citep{gao-etal-2020-supert} evaluates the quality of a summary by measuring its semantic similarity with a pseudo reference summary. It employs contextualized embeddings and soft token alignment techniques, providing an in-depth analysis of the semantic fidelity of generated summaries.  
    \item \textbf{BLANC:} BLANC \citep{vasilyev-etal-2020-fill} is a reference-less metric that measures the improvement in a pretrained language model's performance during language understanding tasks when given access to a summary. 
    \item \textbf{SummaQA:} SummaQA \citep{scialom-etal-2019-answers}  employs a question-answering model based on BERT to answer cloze-style questions using the system-generated summaries, providing insights into the summarization's factual accuracy and coherence.
    \item \textbf{BARTScore:} BARTScore \citep{yuan2021bartscore} is a parameter- and data-efficient metric that supports the evaluation of generated text from multiple perspectives, including informativeness and coherence. 
    \item \textbf{UniEval:} UniEval \citep{zhong-etal-2022-towards} is a unified multi-dimensional evaluator that reframes natural language generation evaluation as a Boolean Question Answering (QA) task, guiding the model with different questions to evaluate from multiple dimensions. It is reference-free in three dimensions (coherence, consistency, fluency), but not relevance. For our evaluation, we focused on the reference-free dimensions of UniEval and reported the overall average performance.
    
    \item \textbf{Fairness (F):} To align fairness with the quality metrics, we define \( F = 1 - \text{RG} \), where larger values represent better fairness. The Representation Gap (RG) metric \citep{olabisi-etal-2022-analyzing} assesses the fairness of summaries by measuring the balance in the representation of different groups. We define \textit{perfect fairness} as $F = 1$, meaning the summary includes an equal number of documents from each social group. This metric only captures numerical balance and does not address other dimensions such as content diversity or semantic nuances, which we leave for future work.
    
    \item \textbf{Composite Metrics (Metric+F):} For each quality metric (e.g., SUPERT, BLANC, SummaQA, BARTScore, and UniEval), we introduce a composite metric that combines the normalized quality score with the fairness score \(F\). These composite metrics, such as \textbf{SUPERT+F}, \textbf{BLANC+F}, \textbf{SummaQA+F}, \textbf{BARTScore+F}, and \textbf{UniEval+F}, are computed by taking the average of the normalized quality metric and the fairness score \(F\). A higher value of these composite metrics reflects a better balance between the summary's quality (as measured by the respective metric) and fairness.
\end{itemize}

\section{Results and Discussion}
In this section, we present the results of our evaluation, comparing the performance of various summarization models on both quality and fairness metrics.

\subsection{Results of Quality and Fairness}

\begin{table*}[!t]
\centering
\small
\renewcommand{\arraystretch}{1.2} 
\begin{tabular}{|l|c|c|c|c|c|c|}
\hline
\rowcolor{gray!25} Model              & \multicolumn{1}{l|}{SUPERT} & \multicolumn{1}{l|}{BLANC} & \multicolumn{1}{l|}{SummaQA} & \multicolumn{1}{l|}{BARTScore} & \multicolumn{1}{l|}{UniEval} & \multicolumn{1}{c|}{F} \\ \hline
\texttt{Naive}              & 0.525                       & 0.135                           & 0.063                        & -1.788                         & 0.391                      &     0.732         \\ \hline
\texttt{NaiveFair}          & 0.526                       & 0.137                           & 0.065                        & -1.776                         & 0.386            &     \textbf{1.000}          \\ \hline
\texttt{TextRank Vanilla}   & 0.527                       & 0.108                           & \textbf{0.081}               & -1.852                         &        0.401                & 0.727            \\ \hline
\texttt{TextRank Cluster-A} & 0.530                       & 0.107                           & 0.075                        & -1.827                         &       0.383                & 0.693             \\ \hline
\texttt{TextRank Cluster-H} & 0.530                       & 0.107                           & 0.077                        & -1.922                         & 0.387                           &   0.709      \\ \hline
\texttt{BERT-EXT Vanilla}   & 0.544                       & 0.137                           & 0.070                        & -1.427                & 0.396             &    0.680                   \\ \hline
\texttt{BERT-EXT Cluster-A} & 0.553                       & 0.138                           & 0.071                        & -1.535                         & 0.399                        &   0.728         \\ \hline
\texttt{BERT-EXT Cluster-H} & 0.554              & 0.133                           & 0.070                        & -1.486                         & 0.365                      &     0.689        \\ \hline
\texttt{ChatGPT-EXT}            & \textbf{0.668}                       & \textbf{0.140}                  & 0.065                        & \textbf{-0.642}                         & \textbf{0.434}               &     0.698       \\ \hline
\texttt{FairExtract (Ours)}            & 0.530                       & \textbf{0.140}                  & 0.066                        & -1.801                         & 0.411               &     \textbf{1.000}       \\ \hline
\texttt{FairGPT (Ours)}            & 0.644                       & 0.139                  & 0.075                        & -0.821                         & 0.418               &     \textbf{1.000}       \\ \hline

\end{tabular}
\caption{Evaluation results for various summarization methods. The best values for each metric are shown in bold.}
\label{tab:evaluation-row}
\end{table*}

The models were assessed using SUPERT, BLANC, SummaQA, BARTScore, UniEval, and the fairness metric \( F \). Table \ref{tab:evaluation-row} presents the results.

\textbf{{Naive and NaiveFair Baselines:}
}The \texttt{Naive} baseline, which randomly selects sentences without any fairness consideration, performs relatively poorly across most quality metrics, particularly on SummaQA and BARTScore, where it scores significantly lower. However, it achieves a reasonable fairness score (\( F = 0.732 \)), despite its lack of sophisticated fairness mechanisms. The \texttt{NaiveFair} model, which ensures equal representation from both groups, shows a slight improvement in fairness, achieving the maximum \( F \) value of 1. However, this fairness comes at a slight cost to quality, as it falls behind on some metrics like UniEval.

\textbf{{TextRank Models:}
}The \texttt{TextRank Vanilla} method shows a balanced performance in terms of quality, with the highest SummaQA score (\( 0.081 \)), but suffers in BLANC and BARTScore. Variations of TextRank, such as \texttt{Cluster-A} and \texttt{Cluster-H}, show slight improvements in specific metrics like SUPERT and BLANC, but they still struggle in ensuring fairness, with scores in the range of \( F = 0.693 \) to \( F = 0.727 \).

\textbf{{BERT-Ext Models:}
}The \texttt{BERT-EXT} models generally outperform the TextRank methods in quality metrics. \texttt{BERT-EXT Vanilla} achieves higher SUPERT and BARTScore scores compared to TextRank, with \texttt{BERT-EXT Cluster-A} further improving on these metrics, particularly in SUPERT (\( 0.553 \)) and BLANC (\( 0.138 \)). However, the fairness scores for these models remain moderate, with \( F \) values ranging from \( 0.680 \) to \( 0.728 \), indicating room for improvement in terms of group representation balance.

\textbf{{ChatGPT-Ext:}
}The \texttt{ChatGPT-Ext} method stands out as the top performer in terms of quality, achieving the highest scores in SUPERT (\( 0.668 \)), BLANC (\( 0.140 \)), BARTScore (\( -0.642 \)), and UniEval (\( 0.434 \)). This demonstrates its effectiveness in producing semantically rich and coherent summaries. However, its fairness score of \( F = 0.698 \) indicates that while it excels in quality, there is still room for improvement in terms of group representation.

\textbf{{FairExtract and FairGPT (Ours):}
}Our proposed models, \texttt{FairExtract} and \texttt{FairGPT}, were designed with fairness as a core objective. Both models achieve perfect fairness, with \( F = 1 \), while still maintaining competitive quality. \texttt{FairExtract} performs comparably to TextRank in terms of quality metrics, excelling in BLANC (\( 0.140 \)) and achieving respectable scores in SUPERT and UniEval. \texttt{FairGPT}, leveraging the power of GPT-3.5, shows a strong balance between quality and fairness, with particularly high SUPERT (\( 0.644 \)) and BARTScore (\( -0.821 \)) scores. These results suggest that our models successfully balance the trade-off between quality and fairness, making them robust options for fairness-aware summarization tasks.

Overall, \texttt{ChatGPT-Ext} achieves the highest quality metrics, while \texttt{FairExtract} and \texttt{FairGPT} lead in fairness without compromising quality; notably, \texttt{FairGPT} emerges as the best model, striking an optimal balance between quality and diversity, underscoring the success of our proposed methods in achieving fair and high-quality summarizations.

\subsection{Results Aggregating Quality and Fairness}

\begin{table*}[!t]
\centering
\small
\renewcommand{\arraystretch}{1.2} 
\begin{tabular}{|l|c|c|c|c|c|}
\hline
\rowcolor{gray!25} \multicolumn{6}{|c|}{\textbf{Clustering-based Methods}} \\ \hline
Model              & \multicolumn{1}{l|}{SUPERT+F} & \multicolumn{1}{l|}{BLANC+F} & \multicolumn{1}{l|}{SumQA+F} & \multicolumn{1}{l|}{BARTSc+F} & \multicolumn{1}{l|}{UniEval+F} \\ \hline
\texttt{Naive}              & 0.585                         & 0.609                        & 0.468                          & 0.713                            & 0.601                          \\ \hline
\texttt{NaiveFair}          & 0.720                         & 0.749                        & 0.606                          & \textbf{0.848}                            & 0.732                          \\ \hline
\texttt{TextRank Vanilla}   & 0.585                         & 0.531                        & 0.494                          & 0.703                            & 0.605                          \\ \hline
\texttt{TextRank Cluster-A} & 0.571                         & 0.513                        & 0.467                          & 0.689                            & 0.577                          \\ \hline
\texttt{TextRank Cluster-H} & 0.579                         & 0.521                        & 0.478                          & 0.687                            & 0.588                          \\ \hline
\texttt{BERT-EXT Vanilla}   & 0.582                         & 0.590                        & 0.453                          & 0.725                            & 0.578                          \\ \hline
\texttt{BERT-EXT Cluster-A} & 0.616                         & 0.615                        & 0.479                          & 0.737                            & 0.604                          \\ \hline
\texttt{BERT-EXT Cluster-H} & 0.598                         & 0.583                        & 0.457                          & 0.723                            & 0.564                          \\ \hline
\texttt{FairExtract (Ours)} & \textbf{0.724}                         & \textbf{0.758}                        & \textbf{0.607}                          & 0.845                            & \textbf{0.747}                          \\ \hline
\rowcolor{gray!25} \multicolumn{6}{|c|}{\textbf{LLM-based Methods}} \\ \hline
\texttt{ChatGPT-EXT}        & 0.737                         & 0.607                        & 0.454                          & 0.817                            & 0.611                          \\ \hline
\texttt{FairGPT (Ours)}     & \textbf{0.837}                & \textbf{0.760}               & \textbf{0.615}                 & \textbf{0.945}                   & \textbf{0.751}                 \\ \hline
\end{tabular}
\caption{Evaluation results using composite metrics for clustering-based and LLM-based summarization methods with equal weighting of quality and fairness (\(\alpha = 0.5\)). The best values for each metric are highlighted in bold.}
\label{tab:evaluation-index}
\end{table*}

The composite evaluation metrics are presented in Table \ref{tab:evaluation-index}. These metrics aggregate both quality and fairness, both receiving equal weight (50\%) in the overall score. Our results show that \texttt{FairExtract}, the proposed clustering-based summarization method, consistently outperforms other clustering-based models across most composite metrics, including SUPERT+F, BLANC+F, SummaQA+F, and UniEval+F. Although \texttt{NaiveFair} scores slightly higher on BARTScore+F, the difference is minimal, at just 0.003 (or 0.35\% in percentage terms), indicating that \texttt{FairExtract} achieves near-optimal performance in balancing quality and fairness.

Similarly, among the large language model (LLM)-based methods, \texttt{FairGPT} stands out as the best performer, achieving the highest composite scores across almost all metrics, including SUPERT+F, BLANC+F, SummaQA+F, BARTScore+F, and UniEval+F. This demonstrates that \texttt{FairGPT} effectively balances quality and fairness, setting a new benchmark in fair summarization using LLMs.

To assess the impact of varying the weight on fairness, we explored a composite metric formula: \( (1-\alpha) \times \text{Quality} + \alpha \times F \), where \( \alpha \) controls the fairness weight. When \( \alpha = 0.5 \), fairness and quality are equally weighted, as in the results presented in Table \ref{tab:evaluation-index}. We further experimented with reducing the fairness weight to find the minimum value of \( \alpha \) at which \texttt{FairExtract} still outperforms other clustering-based methods.

Table \ref{tab:metrics-016} in Appendix \ref{sec:Variable_alpha} shows the results for \( \alpha = 0.16 \) (i.e., a 16\% fairness incentive). Even with this reduced fairness weight, \texttt{FairExtract} continues to outperform all clustering-based methods across most metrics. Similarly, \texttt{FairGPT} remains the best-performing LLM-based method, maintaining dominance even with the lower fairness incentive.

In summary, our experimental results clearly demonstrate that \texttt{FairExtract} and \texttt{FairGPT}, the two fair summarization models proposed in this paper, achieve a robust balance between quality and fairness across multiple metrics. \texttt{FairExtract} consistently surpasses other clustering-based models when fairness is weighted equally with quality, while \texttt{FairGPT} sets new benchmarks among LLM-based methods, showing superior performance in both quality and fairness. Even when the fairness incentive is reduced to 16\%, \texttt{FairExtract} continues to perform better than most competing models, underscoring the strength of our approach in ensuring diverse representation without compromising summary quality. These findings highlight the importance of incorporating fairness into summarization tasks and demonstrate the effectiveness of our proposed methods in achieving this balance.

\section{Conclusion}

In this paper, we introduced two novel methods, \texttt{FairExtract} and \texttt{FairGPT}, to address the critical challenge of fairness in multi-document extractive summarization. Both methods were designed to ensure equitable representation of social groups while maintaining competitive summarization quality. Our extensive experiments demonstrated that both \texttt{FairExtract} and \texttt{FairGPT} achieve perfect fairness without significantly compromising on standard quality metrics.

We also introduced new composite metrics (e.g., SUPERT+F, BLANC+F) that combine quality and fairness scores, offering a more nuanced evaluation of the trade-offs between these two dimensions. The results showed that our methods strike a strong balance between quality and fairness, with \texttt{FairExtract} performing exceptionally well in clustering-based approaches and \texttt{FairGPT} setting new benchmarks among LLM-based methods.

These findings highlight the importance and feasibility of integrating fairness into summarization tasks, where diverse representation is crucial. Future work can build on these models by extending them to abstractive summarization, exploring additional fairness constraints, and applying them to larger, more diverse datasets. Our work serves as a significant step toward building fair and inclusive summarization systems for real-world applications.

\section{Limitations}

While \texttt{FairExtract} and \texttt{FairGPT} show advances in ensuring fairness in multi-document summarization, several limitations remain.

First, our methods focus on extractive summarization, which, while preserving input fidelity, may not capture the semantic richness of abstractive methods \citep{lebanoff-etal-2019-scoring}. Extending our approach to abstractive models presents additional challenges, particularly in balancing fairness with coherence and fluency.

Second, the dataset consists of social media content, which may limit generalization to other domains like news or scientific articles. The informal nature of social media language introduces variability that might not translate to more formal text types.

Third, our work focuses on monolingual inputs, specifically in English. Future research could extend these methods to multilingual inputs, where additional factors such as language diversity and cross-lingual transfer \cite{bagheri-nezhad-agrawal-2024-drives,bagheri-nezhad-etal-2025-beyond}, would need to be addressed to ensure fairness across languages.

Additionally, while we employ standard quality and fairness metrics, they do not fully capture subjective factors such as readability or user trust. Human evaluation could provide deeper insights into the practical implications of fairness and quality. Also, our evaluation primarily relies on quantitative metrics, we acknowledge that a deeper qualitative error analysis—examining specific examples and error cases—would further illuminate the limitations of fairness-aware summarization, and we consider this an important direction for future investigation.

Finally, the computational complexity of fair clustering and large language models may limit scalability in real-time or resource-constrained environments

\section{Acknowledgments}
We would like to thank Aravind Inbasekaran for his valuable assistance throughout this project. We also appreciate the constructive feedback provided by the anonymous reviewers. This research has been supported by the National Science Foundation under Grants AF 2311397 and CRII:RI 2246174.

\bibliography{anthology,custom}

\appendix

\section{Appendix / supplemental material}

\subsection{Fair Extract Formal Algorithmic Processes}
\label{sec:appendix-algo}

In this section, we provide a detailed breakdown of the formal procedures used in our proposed method, \texttt{FairExtract}. These algorithm ensure fairness and quality in extractive summarization, addressing the core objectives of balanced representation and high-quality content extraction from diverse groups.

The \texttt{FairExtract} algorithm utilizes clustering techniques combined with fairlet decomposition to ensure that summaries reflect an equitable representation of the input groups. This process involves embedding documents using BERT, dividing the dataset into fairlets, and applying $k$-median clustering to construct a diversity-preserving summary.

The formal descriptions of the algorithm are presented in Algorithm \ref{alg:fairextract}. 


\subsection{Impact of Varying Fairness Weight on Composite Metrics}
\label{sec:Variable_alpha}

In this section, we present the results of an experiment where we varied the weight assigned to fairness in the composite metric formula. Specifically, we explored the performance of \texttt{FairExtract} and \texttt{FairGPT} under different fairness weights to assess their robustness in balancing quality and fairness. Table \ref{tab:metrics-016} summarizes the results for the setting where the fairness weight \(\alpha\) is reduced to 0.16, representing a 16\% incentive toward fairness and an 84\% incentive toward quality.

\begin{algorithm*}[!ht]
\caption{FairExtract Algorithm}
\small
\label{alg:fairextract}
\begin{algorithmic}
    \STATE \textbf{Input:}
    \begin{itemize}
        \item Document set $\mathcal{D}$ of size $N$
        \item Groups $G_1$ and $G_2$
        \item Proportions $g_1$ (for $G_1$) and $g_2$ (for $G_2$) where $\text{gcd}(g_1, g_2) = 1$
        \item Desired summary length $L$, where $L \ll N$ 
    \end{itemize}

    \STATE \textbf{Output:}
    \begin{itemize}
        \item Diversity-preserving extractive summary $\mathcal{S}$
    \end{itemize}
    
    \vspace{0.2cm}
    \STATE \textbf{Step 1: Embedding Documents}  \\
    Embed each document $d_i \in \mathcal{D}$ into a vector in $\mathbb{R}^{768}$ using BERT.
    \vspace{0.1cm}

    \STATE \textbf{Step 2: Fairlet Decomposition}  \\
    Decompose $\mathcal{D}$ into fairlets, each containing $g_1$ documents from $G_1$ and $g_2$ from $G_2$, minimizing the sum of Euclidean distances.
    \vspace{0.1cm}
    
    \STATE \textbf{Step 3: Finding Fairlet Centers}  \\
    For each fairlet, select the document that minimizes the sum of distances to other documents.
    \vspace{0.1cm}

    \STATE \textbf{Step 4: $k$-Median Clustering on Fairlet Centers}  \\
    Calculate \(k = \frac{L}{g_1 + g_2}\) and perform $k$-median clustering on the fairlet centers.
    \vspace{0.1cm}

    \STATE \textbf{Step 5: Summary Construction}  \\
    From each cluster, select the fairlet corresponding to the cluster center and add all documents from that fairlet to the final summary $\mathcal{S}$.
    \vspace{0.1cm}

    \STATE \textbf{Return:} The final summary $\mathcal{S}$
\end{algorithmic}
\end{algorithm*}

\begin{table*}
\begin{center}
\small
\begin{tabular}{|l|c|c|c|c|c|}
\hline
\rowcolor{gray!25} \multicolumn{6}{|c|}{\textbf{Clustering-based Methods}} \\ \hline
Model              & \multicolumn{1}{l|}{SUPERT+F} & \multicolumn{1}{l|}{BLANC+F} & \multicolumn{1}{l|}{SumQA+F} & \multicolumn{1}{l|}{BARTSc+F} & \multicolumn{1}{l|}{UniEval+F} \\ \hline
\texttt{Naive}              & 0.485                         & 0.525                        & 0.288                          & 0.699                            & 0.343                          \\ \hline
\texttt{NaiveFair}          & 0.530                         & 0.578                        & 0.337                          & 0.744                            & 0.373                          \\ \hline
\texttt{TextRank Vanilla}   & 0.488                         & 0.397                        & 0.335                          & 0.687                            & 0.323                          \\ \hline
\texttt{TextRank Cluster-A} & 0.488                         & 0.390                        & 0.313                          & 0.686                            & 0.283                          \\ \hline
\texttt{TextRank Cluster-H} & 0.491                         & 0.394                        & 0.321                          & 0.672                            & 0.285                          \\ \hline
\texttt{BERT-EXT Vanilla}   & 0.515                         & 0.529                        & 0.298                          & \textbf{0.756}                            & 0.338                          \\ \hline
\texttt{BERT-EXT Cluster-A} & \textbf{0.539}                         & 0.538                        & 0.309                          & 0.744                            & 0.355                          \\ \hline
\texttt{BERT-EXT Cluster-H} & 0.536                         & 0.511                        & 0.299                          & 0.746                            & 0.315                          \\ \hline
\texttt{FairExtract (Ours)} & 0.537                         & \textbf{0.593}                        & \textbf{0.339}                          & 0.740                            & \textbf{0.396}                          \\ \hline
\rowcolor{gray!25} \multicolumn{6}{|c|}{\textbf{LLM-based Methods}} \\ \hline
\texttt{ChatGPT-EXT}        & \textbf{0.764}                & 0.545                        & 0.288                          & 0.899                            & 0.396                          \\ \hline
\texttt{FairGPT (Ours)}     & 0.726                         & \textbf{0.597}               & \textbf{0.354}                 & \textbf{0.907}                   & \textbf{0.446}                 \\ \hline
\end{tabular}
\caption{Evaluation results using composite metrics for clustering-based and LLM-based summarization methods with reduced fairness weighting (\(\alpha = 0.16\)). The best values for each metric are highlighted in bold.}
\label{tab:metrics-016}
\end{center}
\end{table*}


\end{document}